\documentclass{article}

\usepackage{PRIMEarxiv}

\usepackage{mlmath}
\usepackage{booktabs,multirow}   
\usepackage{amsfonts}       
\usepackage{nicefrac}     
\usepackage{microtype}     
\usepackage{xcolor} 
\usepackage{lineno}

\usepackage{times}
\usepackage{graphicx}
\usepackage{subfigure}
\usepackage{caption}
\usepackage{amsmath,amsfonts,amssymb,mathtools,amsthm}
\usepackage{makecell}

\newcommand{\rnull}{\textnormal{null}}
\newcommand{\rnarr}{\textnormal{narr}}
\newcommand{\rwide}{\textnormal{wide}}

\newcommand{\rrank}{\textnormal{rank}}
\newcommand{\rdim}{\textnormal{dim}}
\newcommand{\rspan}{\textnormal{span}}

\newcommand{\TT}{\text{T}}

\newtheorem*{theorem*}{Theorem}
\newtheorem*{definition*}{Definition}

\usepackage[utf8]{inputenc} % allow utf-8 input
\usepackage[T1]{fontenc}    % use 8-bit T1 fonts
\usepackage{hyperref}       % hyperlinks
\usepackage{url}            % simple URL typesetting
\usepackage{booktabs}       % professional-quality tables
\usepackage{amsfonts}       % blackboard math symbols
\usepackage{nicefrac}       % compact symbols for 1/2, etc.
\usepackage{microtype}      % microtypography
\usepackage{lipsum}
\usepackage{fancyhdr}       % header
\usepackage{graphicx}       % graphics
\graphicspath{{media/}}     % organize your images and other figures under media/ folder

\title{Optimistic Estimate Uncovers the Potential of Nonlinear Models
}

\author{
Yaoyu Zhang\textsuperscript{\rm 1,4}\thanks{Corresponding author: zhyy.sjtu@sjtu.edu.cn.},
Zhongwang Zhang\textsuperscript{\rm 1}, 
Leyang Zhang\textsuperscript{\rm 2}, 
Zhiwei Bai\textsuperscript{\rm 1},
Tao Luo\textsuperscript{\rm 1,3},  
Zhi-Qin John Xu\textsuperscript{\rm 1}\thanks{Corresponding author: xuzhiqin@sjtu.edu.cn.}
\\
\textsuperscript{\rm 1}  School of Mathematical Sciences, Institute of Natural Sciences, MOE-LSC \\ 
and Qing Yuan Research Institute, Shanghai Jiao Tong University \\
\textsuperscript{\rm 2} Department of Mathematics, University of Illinois Urbana-Champaign\\
\textsuperscript{\rm 3} CMA-Shanghai\\
\textsuperscript{\rm 4} Shanghai Center for Brain Science and Brain-Inspired Technology\\
}

\begin{document}
\maketitle

\begin{abstract}
We propose an optimistic estimate to evaluate the best possible fitting performance of nonlinear models. It yields an optimistic sample size that quantifies the smallest possible sample size to fit/recover a target function using a nonlinear model. We estimate the optimistic sample sizes for matrix factorization models, deep models, and deep neural networks (DNNs) with fully-connected or convolutional architecture. For each nonlinear model, our estimates predict a specific subset of targets that can be fitted at overparameterization, which are confirmed by our experiments. Our optimistic estimate reveals two special properties of the DNN models---free expressiveness in width and costly expressiveness in connection. These properties suggest the following architecture design principles of DNNs: (i) feel free to add neurons/kernels; (ii) restrain from connecting neurons. Overall, our optimistic estimate theoretically unveils the vast potential of nonlinear models in fitting at overparameterization. Based on this framework, we anticipate gaining a deeper understanding of how and why numerous nonlinear models such as DNNs can effectively realize their potential in practice in the near future.
\end{abstract}

% keywords can be removed

\section{Introduction}
In recent years, large nonlinear models like deep neural networks (DNNs) have achieved unprecedented success, powering amazing applications like AlphaGo \cite{silver2016mastering}, AlphaFold \cite{jumper2021highly}, and GPT \cite{brown2020language,openai2023gpt4}. However, for many decades, the community had severely underestimated their potential. One major factor is that traditional theoretical frameworks for regression and generalization excessively focus on the worst-case scenarios in which overfitting easily happens for overparameterized models. Currently, the faith of the community in DNNs stems from their practical achievements, which are produced in a way analogous to alchemy. The danger lies in the tendency to overestimate the capability of DNNs. This may lead to the misallocation of resources toward unattainable goals, reminiscent of the pursuits of transmuting copper into gold or creating a perpetual motion machine. In this work, to uncover the true potential of nonlinear models like DNNs, we establish a framework of optimistic estimate for nonlinear regression to evaluate their best possible fitting performance.

We demonstrate the idea and significance of an optimistic estimate using a chemical reaction analogy. By applying the law of element conservation, we derive the following optimistic estimates for chemical reactions: graphite has the potential to be converted to diamond, while gold cannot be obtained from copper. These optimistic estimates inform the potential of a chemical reaction, leading to the abandonment of alchemy and subsequently the invention of synthetic diamonds. Likewise, our optimistic estimate for nonlinear regression aims to uncover the potential of nonlinear models in fitting, which could serve as a foundation for their theoretical understanding.

To assess how well a model fits a target function, the sample size necessary for fitting serves as a natural quantity. In linear regression, the sample size is in general determined by the size of model parameters. Therefore, linear models with fewer parameters are preferred in practice. In the case of nonlinear regression, although theoretical understanding is limited, it has been empirically demonstrated that DNN models can effectively fit target functions even at overparameterization \cite{zhang2017understanding}.  For these models, our optimistic estimate provides the smallest possible sample size, referred to as the optimistic sample size, for fitting each target function. It is influenced by both the target function and the model architecture. Moreover,  the optimistic sample size as a functional of the target function exhibits rich structures, especially in the case of DNNs.

Remark that, our theory does not guarantee that a target function $f^*$ with optimistic sample size $R^*$ can be fitted from $R^*$ samples in practice. However, our experiments in matrix factorization models and DNNs show that these models can achieve near-optimism fitting performance with hyperparameter tuning. In this context, ``near-optimism'' means the empirical sample size necessary for fitting $f^*$ in experiments can approach or even match $R^*$. In particular, when $R^*$ is much smaller than the total number of model parameters $M$, the empirical sample size is generally much closer to $R^*$ than to $M$.  The potential mechanism for this phenomenon is discussed in \textit{Conclusions and discussion}.

Our optimistic estimate reveals the vast potential of nonlinear models in fitting at overparameterization. In particular, our estimates uncover the free expressiveness in width and costly expressiveness in connection properties of DNN models. These properties suggest two architecture design principles: (i) feel free to add neurons/kernels; (ii) restrain from connecting neurons. These principles effectively justify the common preference of researchers to enhance the expressive capacity of DNNs by increasing their widths rather than adding connections. Our estimates also suggest quantitative answers to five long-standing problems in nonlinear regression discussed in \textit{Conclusions and Discussion}. These problems include determining which model can fit at overparameterization, the effective size of parameters of a model, the implicit bias of a model, the advantage of neural network models, and the superiority of convolutional architectures. 

\section{Optimistic sample size estimate}
We derive an optimistic estimate for a model $f_{\vtheta}=f(\cdot;\vtheta)$ with $M$ parameters in the following way. In regression, we use a model $f_{\vtheta}$ to fit $n$ data points sampled from a target function $f^*$, aiming to recover $f^*$. Without loss of generality, we consider target functions that can be expressed by model $f_{\vtheta}$, i.e., $f^*\in\fF:=\{f(\cdot;\vtheta)|\vtheta\in\sR^M\}$ \footnote{Otherwise, a larger model should be used for fitting.}. By default, we use gradient descent to solve the regression problem
\begin{equation}
\min_{\vtheta} \frac{1}{n} \sum_{i=1}^n\left(f\left(\vx_i ; \vtheta\right)-f^*(\vx_i)\right)^2.
\end{equation}
The solution is denoted as $f_{\vtheta_{\infty}}$, where $\vtheta_{\infty}$ denotes the model parameters at convergence. In the worst-case scenario, overparameterization prevents us from recovering $f^*$ because of the non-uniqueness of the solution. However, by considering an optimistic scenario with the following best possible initialization, $f^*$ may be recovered from less than $M$ training samples. 
\begin{description}
    \item{\textbf{optimistic initialization:}} initializing in a neighbourhood of an ``optimal'' point $\vtheta^*$ in the target set $\vTheta_{f^*}$.
\end{description}
Here the target set $\vTheta_{f^*}:=\left\{\vtheta \mid f_{\vtheta}=f^*, \vtheta \in \sR^M\right\}$ contains all the parameter points that express the target function.  Since (i) $\vTheta_{f^*}$ may not be a singleton and (ii) initializing near different points in the target set may require different sample sizes for fitting (see the next section for an example), we choose to initialize near an ``optimal'' point in the target set
such that the required sample size attains the minimum.

Above optimistic initialization leads to a best possible fitting scenario. It is possible because any neighbourhood $N(\vtheta^*,\epsilon)\subset\sR^M$ has a positive measure. It is ``best'' because any point in $\vTheta_{f^*}$ recovers $f^*$ with zero generalization error. In contrast, initializing at $\vtheta'\in\vTheta_{f^*}$ is a best but impossible fitting scenario because $\vTheta_{f^*}$ is generally a lower dimensional submanifold of $\sR^M$ with zero measure \cite{cooper2021global}. 

We can analyze the sample size necessary for fitting in a small neighborhood $N(\vtheta',\epsilon)$ with $\vtheta'\in\vTheta_{f^*}$ using the linear approximation $f(\cdot;\vtheta)\approx f^{\mathrm{lin}}\left(\cdot; \vtheta\right) := f^*(\cdot)+\nabla f_{\vtheta'}(\cdot)^\TT(\vtheta-\vtheta')$. Then, the regression problem becomes
\begin{equation}
\min_{\vtheta\in N(\vtheta',\epsilon)} \frac{1}{n} \sum_{i=1}^n\left(f^{\mathrm{lin}}\left(\vx_i ; \vtheta\right)-f^*(\vx_i)\right)^2.
\end{equation}
This is a linear problem with a well-known result: we need $n=\rdim\left(\rspan\left\{\partial_{\theta_i}f(\cdot;\vtheta')\right\}_{i=1}^M\right)$ samples to recover $f^*$. This number serves as an effective number of parameters of the model at $\vtheta'$. Inspired by the rank in differential topology, we refer to this quantity as the model rank defined for any $\vtheta' \in \sR^M$ as
\begin{equation}\label{eq:model_rank_in_parameters}
    R_{f_{\vtheta}}(\vtheta'):=\rdim\left(\rspan\left\{\partial_{\theta_i}f(\cdot;\vtheta')\right\}_{i=1}^M\right).
\end{equation}

Model rank in a nonlinear model can vary across different points in the target set. In the optimistic scenario, the model rank is defined for any function $f'\in\fF$ as the minimum rank among $\vTheta_{f'}$, i.e., $R_{f_{\vtheta}}(f')$ is given by \begin{equation}\label{Eq:rankf}
R_{f_{\vtheta}}(f'):=\min_{\vtheta'\in\vTheta_{f'}}R_{f_{\vtheta}}(\vtheta').
\end{equation}
Note that the notation $R_{f_{\vtheta}}$ is slightly abused for convenience, representing the model rank of either a parameter point or a function depending on its input. Under optimistic initialization, the sample size necessary for fitting $f^*$ is $R_{f_{\vtheta}}(f^*)$. 
We refer to it as the optimistic sample size of $f^*$, which determines the minimum sample size needed to fit $f^*$ using the model $f_{\vtheta}$.

The optimistic sample size exhibits the following properties: (a) $0\leqslant R_{f_{\vtheta}}(f^*)\leqslant M$; and (b) it is influenced by both the model architecture and the target function. Estimating the optimistic sample sizes across $\fF$ allows us to tackle a fundamental problem in nonlinear regression: determining whether a model is capable of fitting at overparameterization. We note that overparameterization is not a well defined notion in machine learning. In particular, naively characterizing the scenario where the sample size $n<M$ as overparameterization results in trivial cases of fitting at overparameterization. For instance, fitting a linear function with the model $f_{\vtheta}=(\theta_1+\theta_2)x$ only need one sample, which is less than $M=2$. In this work, to eliminate such trivial instances, we formally define overparameterization as the scenario of $n<M_{\mathrm{I}}$, where $M_{\mathrm{I}}:=\max_{\vtheta'\in\sR^M}R_{f_{\vtheta}}(\vtheta')$ counts the maximum effective number of parameters of model $f_{\vtheta}$. If the optimistic sample size $R_{f_{\vtheta}}(f')=M_{\mathrm{I}}$ for any $f'\in\fF$, we can infer that model $f_{\vtheta}$ is incapable of fitting at overparameterization. In the subsequent section, we estimate the optimistic sample sizes for two simple models for illustration. We also demonstrate through experiments that the best possible performance estimated under optimistic initialization can be approached under generic random initialization with finely-tuned hyperparameters.   

\section{Optimistic estimate on simple models}\label{sec:rank_analysis_simple}

Initially, we apply our optimistic estimate to a basic linear model $f_{\rm{L}}(\vx;\vtheta)=\theta_0+\theta_1 x_1+\theta_2 x_2$ featuring $M=3$ parameters. Considering any $\vtheta'=[\theta'_0,\theta'_1,\theta'_2]^T\in\sR^3$, we observe that $R_{f_\rL}(\vtheta')=\rdim\left(\rspan\left\{1,x_1,x_2\right\}\right)=3$. Consequently, for any $f'\in\fF_\mathrm{L}:=\{f_{\rm{L}}(\cdot;\vtheta)|\vtheta\in\sR^3\}$, we have $R_{f_\rL}(f')=M_{\mathrm{I}}=3$. Therefore, we conclude that $f_{\rm{L}}$ is unable to fit at overparameterization. Analogously, we can demonstrate that every linear model possesses a fixed optimistic sample size. As a result, all linear models are unable to fit at overparameterization. This finding aligns with the principles of linear regression theory.

The optimistic estimate is subject to change when we make a slight reparameterization of the linear model as $f_{\mathrm{NL}}(\vx;\vtheta) = \theta_0 + \theta_1 x_1 + \theta_2 \theta_3 x_2$. While the function space remains the same as $\fF_\mathrm{NL} = \fF_\mathrm{L}$, the model becomes nonlinear in terms of its parameters. The optimistic sample size of the model is estimated as follows. Model rank as a function over the parameter space $\sR^4$ is obtained as
\begin{equation}
R_{f_\mathrm{NL}}(\vtheta') = \rdim\left(\rspan\{1, x_1, \theta_3'x_2, \theta_2'x_2\}\right) =\begin{cases}
2, & \theta'_2=\theta'_3=0, \\
3, & \textrm{others}.
\end{cases} \label{def:rank_theta}
\end{equation}
Therefore, we have $M_{\mathrm{I}} = 3$. By solving the minimization problem Eq. \eqref{Eq:rankf}, we obtain the model rank as a function over the function space $\fF$ as
\begin{equation}\label{eq:rank_NL}
R_{f_\mathrm{NL}}(f^*) = \begin{cases}
2, & f^*\in\{a_0+a_1x_1|a_0,a_1\in\sR\}, \\
3, & f^*\in\{a_0+a_1x_1+a_2x_2|a_2 \neq 0, a_0,a_1,a_2\in\sR\}.
\end{cases}
\end{equation}
Above estimate indicates that the nonlinear model $f_{\mathrm{NL}}$ can potentially fit $f^*\in\{a_0+a_1x_1|a_0,a_1\in\sR\}$ at overparameterization using two samples, but it cannot achieve the same for $f^*\in\{a_0+a_1x_1+a_2x_2|a_2 \neq 0, a_0,a_1,a_2\in\sR\}$.

\begin{figure}[htbp]
	\centering
 	\subfigure[$f_{\rL}(\vx;\vtheta)$]{\includegraphics[width=0.43\textwidth]{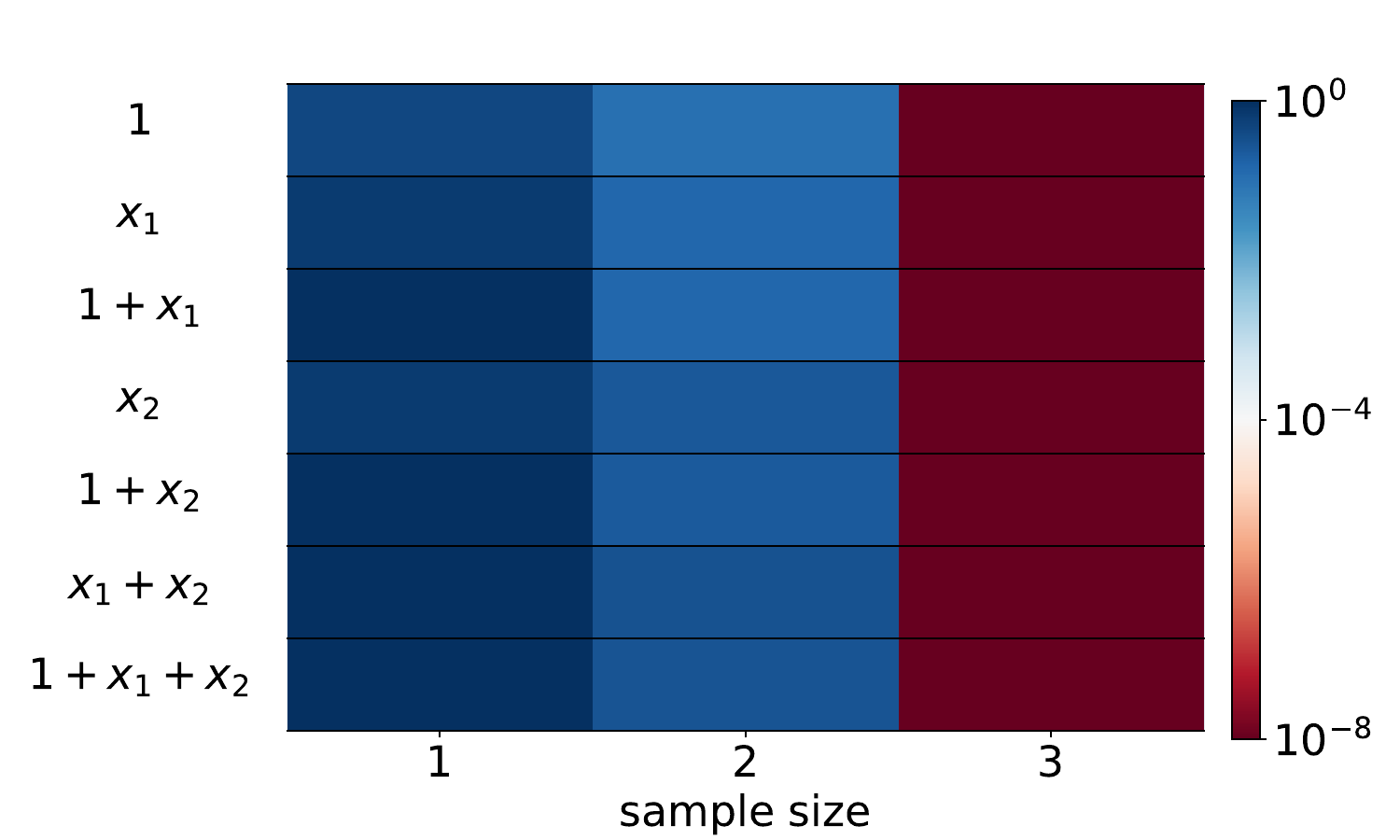}}
	\subfigure[$f_{\mathrm{NL}}(\vx;\vtheta)$]{\includegraphics[width=0.43\textwidth]{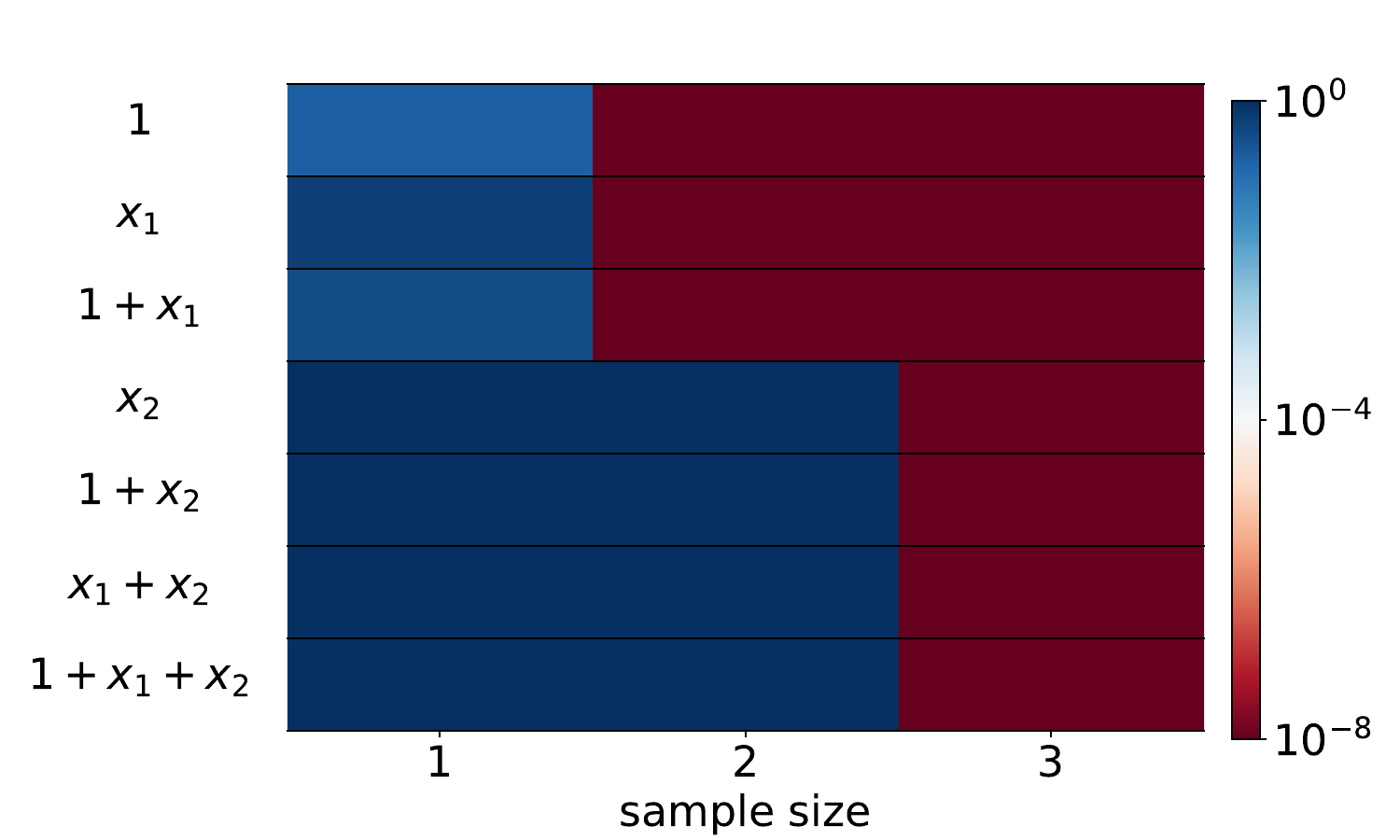}}

  \caption{Average test error (color) vs. the number of training samples (abscissa) for different target functions (ordinate).  Each test error is averaged over $100$ trials with random initialization.} \label{fig:random_feature_bar}
\end{figure}

It is important to note that our theory does not provide a guarantee that the optimistic fitting performance can be achieved in practice. Nevertheless, our experiments demonstrate that, when hyperparameters are well-tuned, the fitting performance of numerous models can closely approximate, or even match, our optimistic estimate. For instance, in Fig. \ref{fig:random_feature_bar}, we utilize gradient descent with a small initialization, wherein model parameters are randomly initialized with a small variance, to fit one, two, or three data points randomly sampled from target functions labeled on the ordinate. The color red represents a fitting (generalization) error that is close to zero. In the case of the linear model $f_{\mathrm{L}}$, every target function requires three samples to achieve almost $0$ fitting/generalization error. Conversely, for the nonlinear model $f_{\mathrm{NL}}$, a mere two samples are sufficient to fit target functions $1$, $x_1$, and $1+x_1$. Evidently, these simple models can achieve the optimistic fitting performance indicated by our optimistic estimate.

The above optimistic estimate can be easily extended to deep models with application in compressed sensing \cite{woodworth2020kernel}. For example, a simple depth-$2$ model reads 
\begin{equation}\label{eq:depth2_model}
    f_{\vtheta}(\vx)=\sum_{j=1}^{d} \left(a_j^2-b_j^2\right) x_j,
\end{equation}
with $\vtheta=(\{a_j\}_{j=1}^{d},\{b_j\}_{j=1}^{d})$ and $\vx = [x_1,\cdots,x_d]^\TT$. For any target function $f^*$ with $k$ nonzero coefficients in its expression (i.e., it is $k$-sparse), its optimistic sample size is
\begin{equation}
    R_{f_{\vtheta}}(f^*)=k.
\end{equation}
In accordance with \cite{woodworth2020kernel}, this model with small initialization implicitly minimizes $\sum_{j=1}^{d}|\gamma_j|$, where $\gamma_j=a_j^2-b_j^2$. By the theory of compressed sensing \cite{candes2008introduction}, this model can fit a $k$-sparse target function generally from $O(k\log d)$ samples, which is much closer to $O(k)$ than to $O(d)$ for sparse targets with $k \ll d$.

It is observed that deep models of various forms, such as model
\begin{equation}\label{eq:deep_model}
    f_{\vtheta}(\vx)=\sum_{j=1}^{d} \left(a_j^{[1]}\cdots a_j^{[L]}\right) x_j
\end{equation}
with $L\geq 2$ and $\vtheta=(\{a_j^{[1]}\}_{j=1}^{d},\cdots,\{a_j^{[L]}\}_{j=1}^{d})$, empirically exhibit a similar implicit bias toward sparse interpolations when trained with small initialization. However, extending the analysis of implicit bias presented in Ref. \cite{woodworth2020kernel} to these models is challenging. In contrast, our optimistic estimate can be easily applied to these models. For model Eq. \eqref{eq:deep_model}, regardless of the depth $L$, the optimistic sample size of a $k$-sparse target function remains $k$. This finding indicates that model Eq. \eqref{eq:deep_model} and model Eq. \eqref{eq:depth2_model} belong to the same class of optimistically equivalent models, suggesting they possess the same potential for fitting at overparameterization. Generally, it is reasonable to expect models within the same optimistically equivalent class to exhibit similar implicit bias and fitting performance. 

In the following sections, we will present the optimistic sample sizes for several widely adopted nonlinear models, including matrix factorization models, fully-connected and convolutional NNs. Our focus lies in: (i) understanding the impact of both the model architecture and the target function on the optimistic sample size, and (ii) investigating the proximity between empirical sample sizes obtained in experiments and their corresponding optimistic sample sizes.

\section{Optimistic estimate on matrix factorization models}
In this section, we apply our optimistic estimate to a nonlinear matrix factorization model denoted as $\vf_{\vtheta} = \mA\mB$. This model finds application in matrix completion tasks, where the objective is to recover a target matrix $\mM^*$ from $n$ observed entries $S=\left\{\left((i_s,j_s),\mM^*_{i_{s}j_{s}}\right)\right\}_{s=1}^n$ \cite{srebro2004learning}. To solve this problem, we employ gradient descent and minimize the following objective:
\begin{equation}
    \min_{\vtheta} \frac{1}{n} \sum_{s=1}^n\left([\vf_{\vtheta}]_{i_{s}j_{s}}-\mM^*_{i_{s}j_{s}}\right)^2,
\end{equation}
where $\mM^*\in\sR^{d\times d}$, $\vtheta=(\mA,\mB)$, and $\mA,\mB\in\sR^{d\times d}$. By employing a standard optimistic estimate procedure (see Appendix for details), we determine the optimistic sample size for any $\mM^*\in\sR^{d\times d}$ as:
\begin{equation}\label{eq:rank_MF}
    R_{\vf_{\vtheta}}(\mM^*)=2r_{\mM^*}d-r_{\mM^*}^2,
\end{equation}
where $r_{\mM^*}=\rrank(\mM^*)$ is the matrix rank of $\mM^*$. Notably, for a fixed $d$, the optimistic sample size increases with $r_{\mM^*}$. Particularly, for large $d$, the optimistic sample size for a rank-$o(d)$ target matrix is $o(d^2)$, significantly smaller than $M_{\rI}=d^2$. Consequently, the matrix factorization model can potentially fit low rank matrices at overparameterization. This finding aligns with existing researches on matrix factorization models, which suggests an implicit bias towards lower rank completions \cite{gunasekar2017implicit,arora2019implicit}.

\begin{figure}[htbp]
	\centering
	\includegraphics[width=0.6\textwidth]{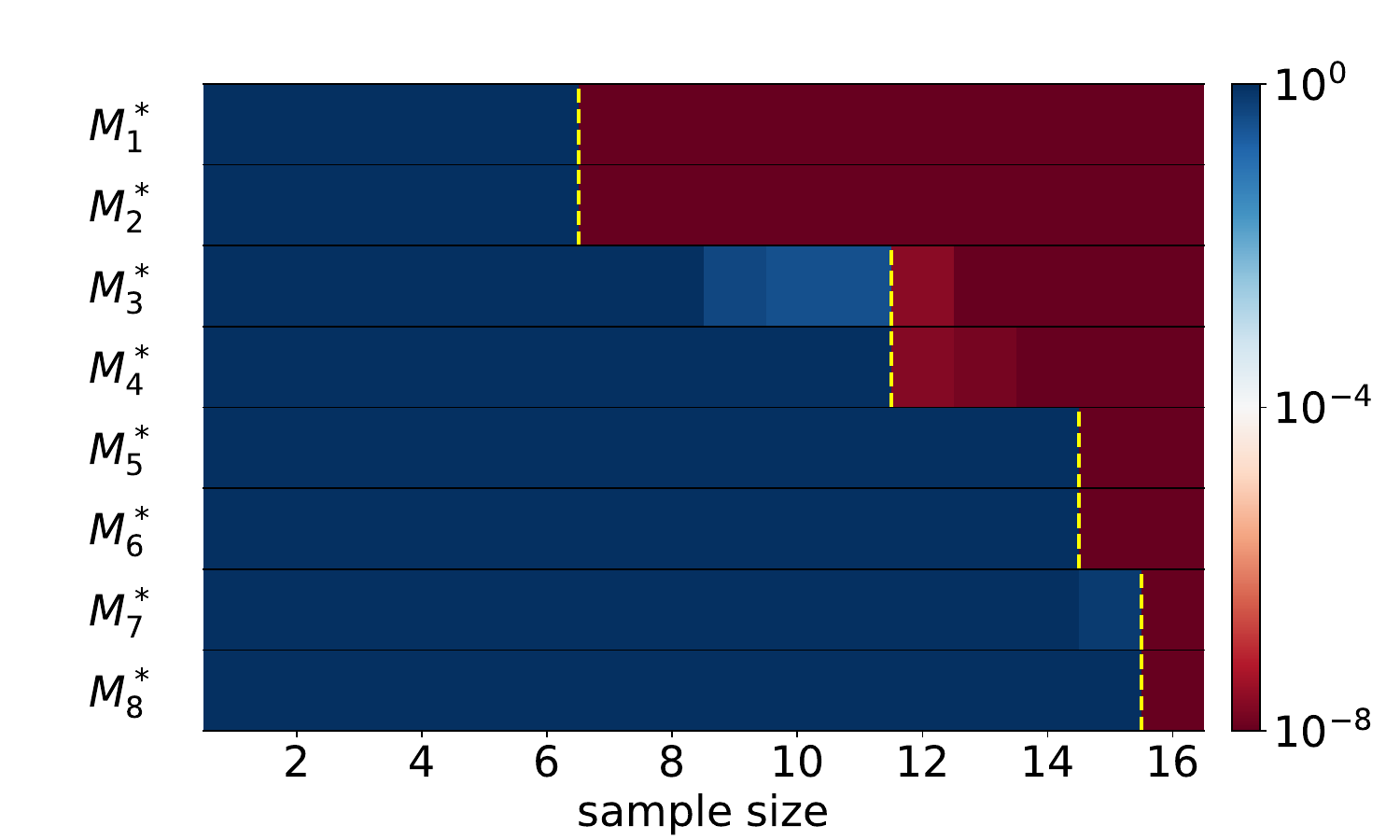}
	
  \caption{Average test error (color) vs. the number of training samples (abscissa) for different target functions (ordinate). The yellow dashed lines indicate the transitions to the corresponding optimistic sample sizes. Different target matrices are considered in different rows, where $\rrank(\mM_{2k-1}^*)=\rrank(\mM_{2k}^*)=k$ for $k=1,2,3,4$. Each test error is averaged over $50$ trials with random initialization.} \label{fig:matrix_com}
\end{figure}

We also conducted experimental tests to demonstrate the practical achievability of the optimistic sample sizes derived above. Referring to equation Eq. \eqref{eq:rank_MF}, a $4\times 4$ target matrix $\mM^*$ with a rank of $1$, $2$, $3$, or $4$ corresponds to an optimistic sample size of $7$, $12$, $15$, or $16$, respectively. Fig. \ref{fig:matrix_com} illustrates the fitting error of eight target matrices using sample sizes ranging from $1$ to $16$. The empirical sample size for fitting corresponds to the point at which the fitting/generalization error approaches zero for the first time. As depicted in Fig. \ref{fig:matrix_com}, when employing a random initialization with a small variance, the empirical sample sizes align with the corresponding optimistic sample sizes indicated by the yellow dashes.

\subsection{Role of hyperparameter tuning}
\begin{figure}[htbp]
	\centering
 	\subfigure[use $f_{\rm NL}$ to fit, $1+x_1$]{\includegraphics[width=0.43\textwidth]{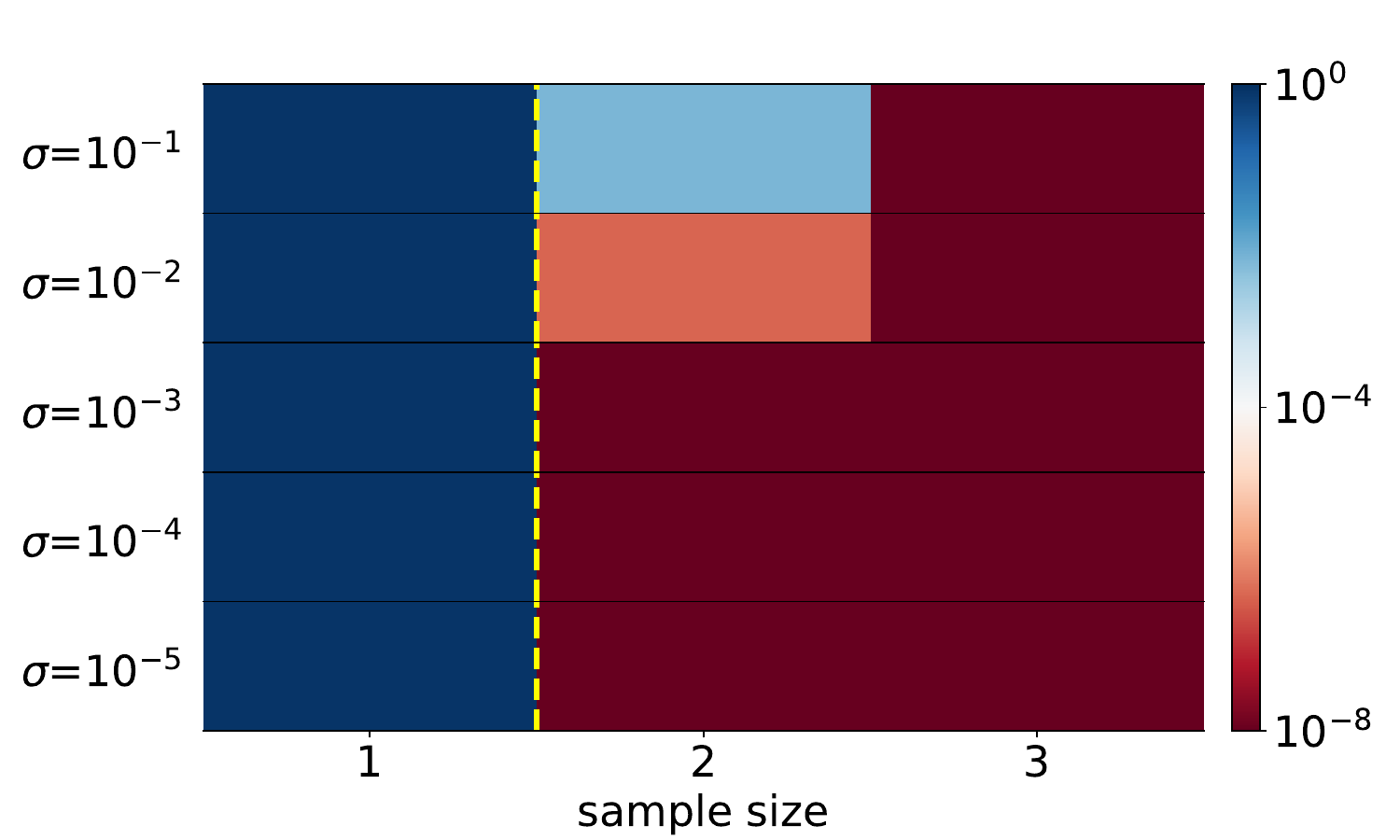}}
	\subfigure[matrix completion, rank=1]{\includegraphics[width=0.43\textwidth]{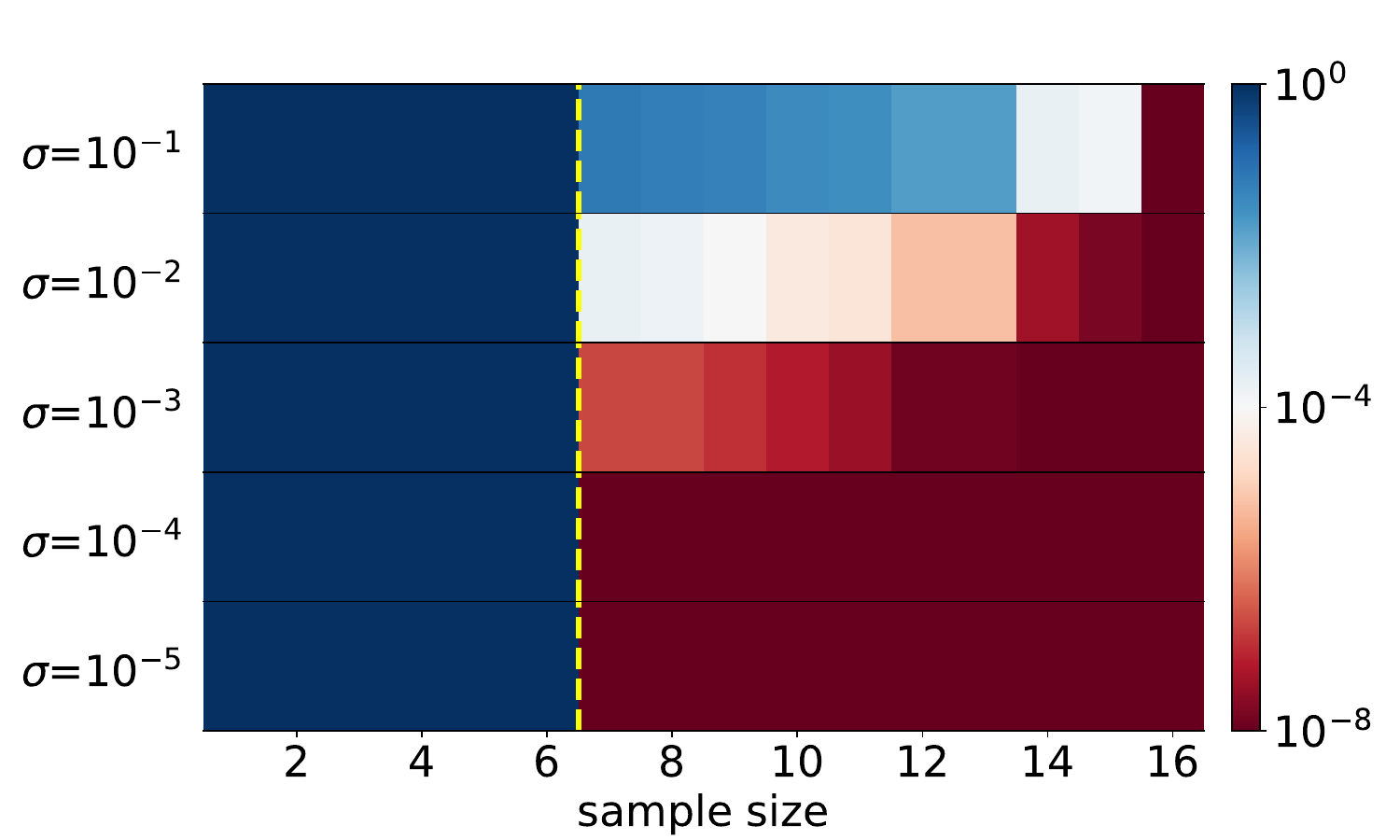}}
\caption{Average test error (color) vs. the number of training samples (abscissa) over different sizes of standard deviation of initialization (ordinate). (a) Performance of fitting $1+x$ by $f_{\rm NL}(x,\theta)$. (b) Performance of completing a rank-$1$ matrix by the matrix factorization model. The yellow dashed lines indicate the transitions to the optimistic sample sizes. Each test error is averaged over $50$ trials with random initialization. } \label{fig:diff_ini}
\end{figure}

Incorrect configuration of training hyperparameters, exemplified by using random initialization with a high variance as depicted in Fig. \ref{fig:diff_ini}, can lead to empirical sample sizes significantly larger than the optimistic ones. As shown in Fig. \ref{fig:diff_ini}, tuning the variance at initialization to a small value is critical for the matrix factorization model to achieve near-optimism fitting performance. Alternatively, oversampling the training data beyond the optimistic sample size alleviates the necessity for extensive fine-tuning. These two characteristics corroborate the widely observed phenomenon in practical applications of large nonlinear models, wherein hyperparameter tuning and a substantial training dataset significantly augment their fitting performance at overparameterization.

\section{Optimistic estimate on DNNs}\label{sec:DNNs}
The DNN family stands out as the most important group of nonlinear models for our optimistic estimate. Nevertheless, determining their optimistic sample sizes proves to be considerably more challenging compared to the aforementioned models. In this section, we present exact results for two-layer tanh-NNs with fully-connected or convolutional architectures. Additionally, we demonstrate via experiments that the empirical sample sizes can approach the optimistic sample sizes for DNNs. To address deeper NNs, we establish an upper bound on their optimistic sample sizes.

\subsection{Optimistic sample sizes of two-layer fully-connected and convolutional NNs}
We first consider a two-layer fully-connected neural network with hyperbolic tangent activation function (tanh-NN) with $m$ hidden neurons, represented by $f_{\boldsymbol{\theta}}(\boldsymbol{x}) = \sum_{i=1}^{m}a_i\tanh(\boldsymbol{w}_i^\mathrm{T} \boldsymbol{x})$, where $\boldsymbol{x}\in \mathbb{R}^d, \boldsymbol{\theta} = (a_i\in\sR, \boldsymbol{w}_i\in\sR^d)_{i=1}^m$. This network has a total of $M=m(d+1)$ parameters. By our optimistic estimate (see Appendix for details), the optimistic sample size is a function of the intrinsic width of the target function $k(f^*)$. Here, the intrinsic width $k(f^*)$ is defined as the minimum width of an NN that can express $f^*$, i.e., $f^*$ can be expressed by a width-$k(f^*)$ NN but not any narrower NN. For any $f^*$ expressible by the width-$m$ NN, its intrinsic width $0\leqslant k(f^*)\leqslant m$. The optimistic sample size of $f^*$ is
\begin{equation}\label{eq:rank_NN}
    R_{\mathrm{NN}_m}(f^*)= k(f^*)(d+1),
\end{equation}
which increases linearly with the intrinsic width. Therefore, a function with intrinsic width $k(f^*)\ll m$ can be fitted optimistically at heavy overparameterization.  

We then consider a two-layer tanh-CNN with weight-sharing represented by
\begin{align}\label{eq:tanh-CNN_simp}
    f_{\vtheta} (\vx) = \sum_{i=1}^{m_{\mathrm{C}}} \sum_{j=1}^{d+1-s} a_{ij} \tanh\left(\sum_{\alpha=1}^{s} x_{j+s-\alpha} K_{i;\alpha}\right),
\end{align}
with $\vx=[x_1,\cdots,x_d]^\TT\in \sR^{d}$.
This CNN with $1$-D convolution has $M={m_{\mathrm{C}}}(d+1)$ free parameters, where ${m_{\mathrm{C}}}$ is the number of convolution kernels analogous to the width of a fully-connected NN. By our optimistic estimate (see Appendix for details), similar to the fully-connected case, the optimistic sample size is a function of the intrinsic number of kernels $k_{\rC}(f^*)$ of the target function $f^*$. Specifically, for any $f^*$ expressible by the $m_{\rC}$-kernel CNN, its optimistic sample size is
\begin{equation}\label{eq:rank_CNN}
    R_{\mathrm{CNN}_{m_{\mathrm{C}}}}(f^*)= k_{\rC}(f^*)(d+1).
\end{equation}
Therefore, a function with intrinsic number of kernel $k_{\rC}(f^*)\ll{m_{\mathrm{C}}}$ can be fitted optimistically at heavy overparameterization.

\subsection{Optimistic properties of two-layer NNs}

Our optimistic estimate provides a means to theoretically quantifies the best possible performance of different models in fitting a target function. It is useful for model selection and architecture design. For example, given a target function $f^*$, we can trace the change of its optimistic sample size with the design of model architecture to study how model architecture influences the fitting performance.

We first use this approach to study the impact of width for two-layer tanh-NNs. Given a target function with intrinsic width $k$, it can be expressed by all NNs with width $m\geqslant k$. According to Eq. \eqref{eq:rank_NN}, all of them yield the same optimistic sample size $k(d+1)$. Therefore, their best possible fitting performance remains the same. A similar result applies to two-layer tanh-CNNs, in which a target function with intrinsic number of kernels $k$ can be fitted optimistically from $k(d+1)$ samples for all CNNs with $m\geqslant k$ kernels. Clearly, for both two-layer fully-connected and convolutional NNs, an increase in the width (or number of kernels) enhances their expressiveness without impairing the optimistic fitting performance.
We call this property \textit{free expressiveness in width}. This property appears to conflict with the Occam Razor principle. However, it is worth noting that the Occam Razor accounts for the worst-case scenario, while the free expressiveness property accounts for the best possible scenario. As demonstrated by our experiments, performance estimated at the best possible scenario is often close to the empiricial performance especially with finely tuned hyperparameters. 

We then investigate how interlayer connectivity between neurons affects the fitting performance. For instance, a two-layer fully-connected neural network connects every input neuron to a hidden neuron, while a two-layer CNN connects only a local group of input neurons to a hidden neuron. Let $f^*$ be a target function with intrinsic number of kernels $k$ (each kernel consist of $d+1-s$ neurons). For the convenience of discussion, we further assume the intrinsic width of $f^*$ is $k(d+1-s)$. Then, for all CNNs with $m_{\mathrm{C}}>k$ kernels, $R_{\mathrm{CNN}_{m_{\rC}}}(f^*)=k(d+1)$. For all fully-connected NNs with $m>k(d+1-s)$ neurons, $R_{\mathrm{NN}_{m_{\rC}}}(f^*)=k(d + 1 - s)(d+1)$. In practice, because input dimension $d$ is often much larger than the convolution kernel size $s$,  we have $k(d+1)\ll k(d + 1 - s)(d+1)$ indicating that CNNs are superior to fully-connected NNs regarding the CNN functions.

Note that both reduced connectivity and weight-sharing contribute to the smaller optimistic sample sizes of CNNs. To identify the contribution purely from the reduced connectivity between neurons, we also  estimate the optimistic sample size for a no-sharing CNN (see Appendix for details). For all no-sharing CNNs with kernel number $m_\mathrm{C} \geqslant k$, the optimistic sample size is $R_{\mathrm{CNN}^{\mathrm{NS}}_{m_\mathrm{C}}}(f^*)=k(s + 1)(d + 1 - s)$. Therefore, increasing the number of connections per hidden neuron from $s$ to $d$ multiplies the optimistic sample size by $\frac{d+1}{s+1}$. To illustrate the difference, we consider a target function expressed by a $k$-kernel CNN with $3\times 3$ convolution kernel and a $28\times 28$ input. In this scenario, the optimistic sample size is $685k$ for a CNN with $3\times 3$ convolution kernels, $6760k$ when weight sharing is removed, $530660k$ when all neurons in adjacent layers are connected to one another \footnote{Here, we consider CNNs with 2-d convolution, whose optimistic sample size is slightly different from that for 1-d convolution (see Appendix for details).}. Apparently, unnecessary connections between neurons increase the NN's expressiveness at a high cost of sample efficiency. We call this property  \textit{costly expressiveness in connection}.

In general, we refer to the properties of models obtained from optimistic estimates as their optimistic properties. Above, we uncover two optimistic properties of two-layer tanh-NNs---free expressiveness in width and costly expressiveness in connection. These properties suggest two architecture design principles for NNs: (i) feel free to add neurons/kernels, and (ii) restrain from connecting neurons. These two principles justifies the common practice that one would rather expanding the width (or number of kernels) than adding connections to increase the expressiveness of NNs.

\subsection{Numerical experiments}

\begin{figure}[htbp]
	\centering
	\subfigure[fully-connected NN]{\includegraphics[width=0.44\textwidth]{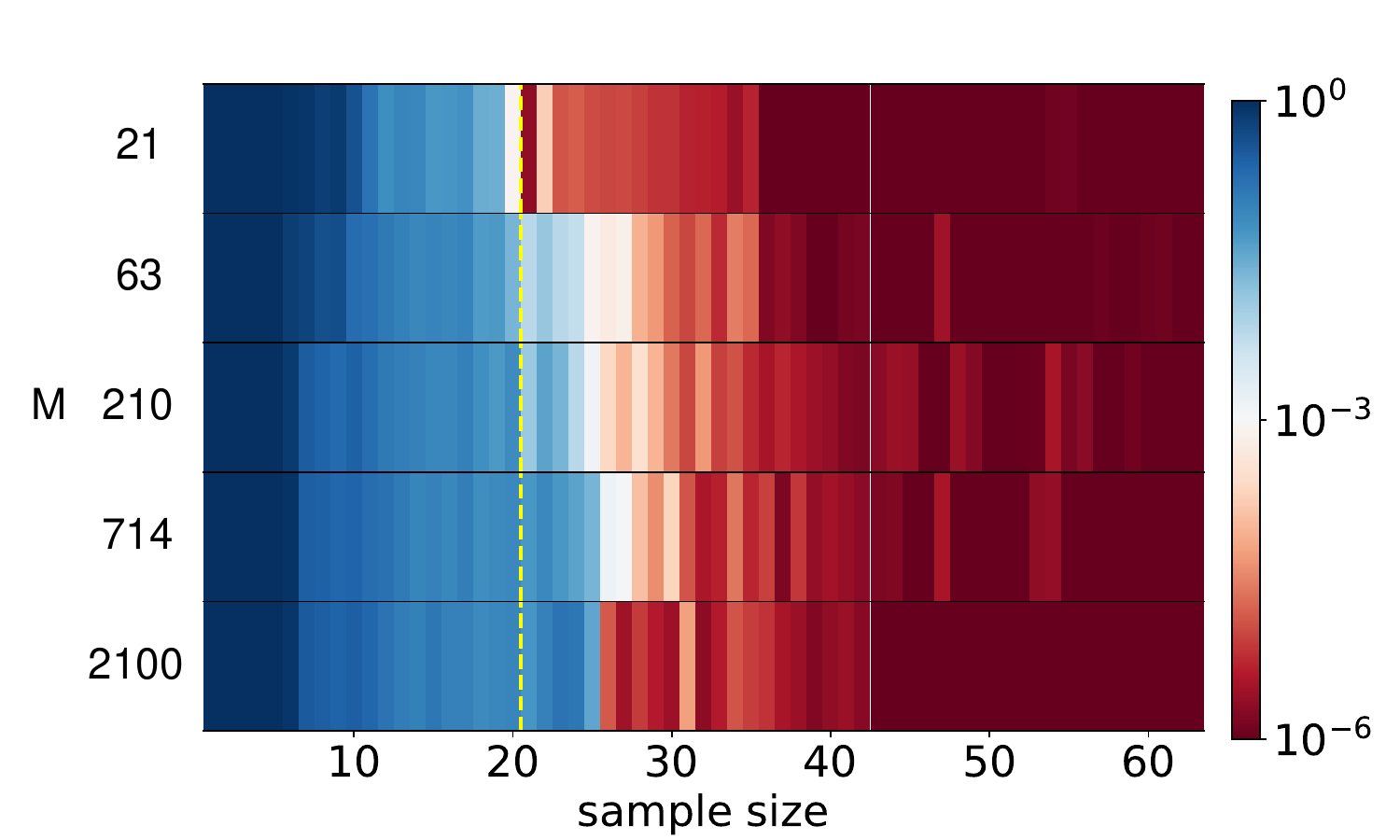}}
        \subfigure[CNN]
    {\includegraphics[width=0.44\textwidth]{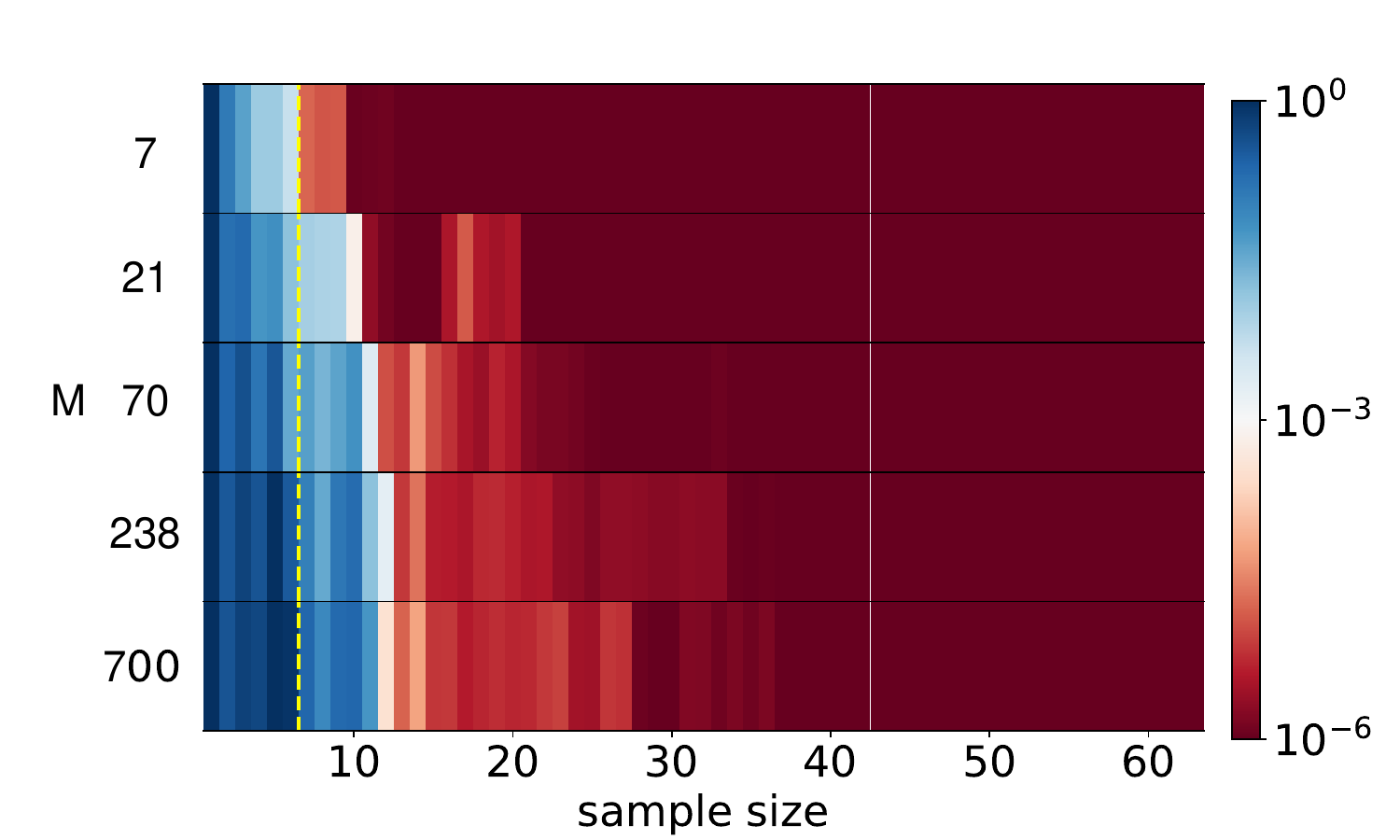}}

\caption{\label{fig: network_stab}Average test error (color) for NNs with different parameter sizes (ordinate) over different training sample sizes (abscissa) in fitting the target function Eq. \eqref{eq:NN_target}. Results are exhibited for (a) two-layer fully-connected tanh-NN (with bias) and (b) two-layer tanh-CNN (with bias). The yellow dashed lines indicate the transitions to the corresponding optimistic sample sizes. For each experiment, the training hyperparameters are finely tuned (see Appendix for details).
} 

\end{figure}

Fig. \ref{fig: network_stab} presents an analysis of empirical sample sizes with fine tuning for two-layer tanh-NNs with different widths and architectures. The target function used in our experiments reads
\begin{equation}\label{eq:NN_target}
    f^{*}(\boldsymbol{x}) = \sum_{s=1}^{3}\tanh(0.6x_{s}+0.8x_{s+1}+x_{s+2}).
\end{equation}
We train two-layer fully-connected and convolutional tanh-NNs with various widths (parameter sizes) on randomly sampled training datasets of various sizes from $1$ to $63$. Remark that, due to the free expressiveness in width, the optimistic sample size indicated by the yellow dashed line does not increase as the size of NN expands. As shown in Fig. \ref{fig: network_stab}, for both architectures, the empirical sample sizes are close to the corresponding optimistic sample sizes over a wide range of parameter sizes. In contrast, they are often far less than the parameter sizes especially for large NNs. These results demonstrate that our optimistic estimate accounting for the best possible scenario is highly relevant to the practical performance of NNs with hyperparameter tuning. The potential mechanism underlying the near-optimism fitting performance of NNs is discussed in \textit{Conclusions and discussion}.

\subsection{Free expressiveness in width of general DNNs}
Estimating $R_{f_{\vtheta}}(\vtheta)$ across the parameter space for a general DNN model presents a challenge. Additionally, identifying the minimum model rank within a target set $\vTheta_{f^*}$ is difficult due to its intricate geometry. The exact estimate of optimistic sample sizes for general DNNs is left for future study. Nevertheless, by employing the powerful tool of critical embeddings \cite{zhang2021embedding,zhang2022embedding}, we can derive an upper bound estimate of the optimistic sample size (see Appendix for details):
\begin{equation}
    R_{\mathrm{NN}_\rwide}(f^*)\leqslant R_{\mathrm{NN}_\rnarr}(f^*)\leqslant M_{\rnarr},
\end{equation}
for any $f^*\in\fF_{\mathrm{NN}_\rnarr}\subset \fF_{\mathrm{NN}_\rwide}$, where $\mathrm{NN}_\rwide$ denotes a wider DNN with no fewer neurons in each layer than $\mathrm{NN}_\rnarr$.

The above upper bound estimate establishes the property of free expressiveness in width for general DNNs, which states:
\begin{description}
    \item{\textbf{Free expressiveness in width:}} Expanding the width of a DNN does not increase the optimistic sample size.
\end{description}
Consequently, the optimistic sample size for a target function $f^*$ within a wide DNN does not exceed the size of the minimum-width DNN capable of expressing $f^*$. Hence, the optimistic sample size for a general DNN adapts to the intrinsic complexity of the target function, rather than the complexity of the DNN model employed for fitting.

\section{Conclusions and discussion\label{sec:conclusion}}
In this work, we establish an optimistic estimate framework for nonlinear regression, which uncovers the best possible fitting performance of nonlinear models at overparameterization. We apply this framework to matrix factorization models, deep models and NNs with fully-connected or convolutional architectures. Our results predict the target functions that these models can fit at overparameterization, which have been empirically confirmed through experiments.

\subsection{Optimistic estimate sheds light to long-standing problems}
Based on our estimates and experimental findings, we propose an optimistic hypothesis that many nonlinear models, including deep neural networks (DNNs), can achieve near-optimism fitting performance. However, to validate and support this hypothesis, further comprehensive experiments and theoretical studies are required. To inspire further research in this direction, we note that our optimistic estimate suggests quantitative answers to five long-standing problems in nonlinear regression as follows.
\\
\textit{(1) Models capable of fitting at overparameterization:} nonlinear models characterized by non-constant optimistic sample sizes. \\
\textit{(2) Effective size of parameters of a nonlinear model $f_{\vtheta}$\footnote{This problem had been proposed by Leo Breiman specifically for NNs almost three decades ago \cite{breiman1995reflections}.}:} $R_{f_{\vtheta}}(f)$ for any target function $f\in\fF$.\\
\textit{(3) Implicit bias of a nonlinear model:} biased towards interpolations with lower optimistic sample sizes.\\
\textit{(4) Advantage of NN models:} free expressiveness in width, i.e., expressiveness of a NN can be improved through widening without increasing the optimistic sample size.\\
\textit{(5) Superiority of the local connection in CNNs:} costly expressiveness in connection, i.e., removing unnecessary connections between neurons substantially reduces the optimistic sample size.\\

\subsection{Potential mechanism to the near-optimism performance}
The near-optimism fitting performance of DNNs is closely related to the condensation phenomenon \cite{luo2021phase,zhou2022towards,zhou2022empirical} observed during the nonlinear training process beyond the neural tangent kernel regime \cite{jacot2018neural}. Specifically, in a two-layer ReLU-NN, condensation happens when the input weights of hidden neurons become concentrated on isolated orientations. The model rank of a condensed ReLU network is significantly lower than the number of network parameters. Ref. \cite{luo2021phase} demonstrated that condensation happens commonly  in the nonlinear regime of the phase diagram for both synthetic and real datasets for wide two-layer ReLU networks. Similar observations have been reported for three-layer ReLU NNs \cite{zhou2022empirical} and networks employing different activation functions \cite{zhou2022towards}. Additionally, it has been shown that condensation can be facilitated by employing large learning rates \cite{andriushchenko2022sgd} and dropout  techniques \cite{zhang2022implicit}. Theoretical research regarding condensation has provided insights into its occurrence during the initial stage of training \cite{maennel2018gradient,pellegrini2020analytic,zhou2022towards}. In future investigations, we aim to delve deeper into the role of condensation in enabling DNNs to achieve near-optimism fitting performance.

\subsection{Implicit bias of nonlinear models}
Investigating the implicit bias present at overparameterization constitutes a significant challenge in the fields of deep learning and nonlinear regression. Our optimistic estimate reveals an inherent bias towards interpolations characterized by lower optimistic sample sizes, as elaborated below. When provided with $n$ training samples, a nonlinear model $f_{\vtheta}$ tends to acquire an interpolation $f'$ with an optimistic sample size $R_{f_{\vtheta}}(f')\leqslant n$, which may not match the target function. For instance, in a matrix completion problem with $2rd-r^2$ observed entries, a matrix factorization model tends to find a completion characterized by a matrix rank no more than $r$. This implicit bias highlights the inherent adaptability of nonlinear models, wherein the complexity of the acquired solution adjusts to match the complexity of the training dataset. As suggested by our prior studies \cite{zhou2022towards,zhang2021embedding}, this bias may be implicitly attained through an adaptive training process during which the model's complexity, measured by the model rank, gradually increases. Further elaborations regarding this implicit bias will be provided in our subsequent research works.

\section*{Acknowledgments}
This work is sponsored by the National Key R\&D Program of China  Grant No. 2022YFA1008200 (Z. X., T. L., Y. Z.), the National Natural Science Foundation of China Grant No. 12101402 (Y. Z.), No. 62002221 (Z. X.), No. 12101401 (T. L.), the Lingang Laboratory Grant No.LG-QS-202202-08 (Y. Z.), Shanghai Municipal of Science and Technology Project Grant No. 20JC1419500 (Y. Z.), the Shanghai Sailing Program (Z. X.), the Natural Science Foundation of Shanghai Grant No. 20ZR1429000  (Z. X.), Shanghai Municipal Science and Technology Key Project No. 22JC1401500 (T. L.), Shanghai Municipal of Science and Technology Major Project No. 2021SHZDZX0102, and the HPC of School of Mathematical Sciences and the Student Innovation Center at Shanghai Jiao Tong University. We also thank the insightful suggestions from Lei Wu.

%Bibliography
\bibliographystyle{unsrt}  
\bibliography{references}

\appendix

\section{Estimate of the optimistic sample sizes}

\begin{definition*}[model rank]
Given any differentiable (in parameters) model $f_{\vtheta}$, the model rank for any $\vtheta^*\in \sR^M$ is defined as
\begin{equation}
R_{f_{\vtheta}}(\vtheta^*):=\rdim\left(\rspan\left\{\partial_{\theta_i} f(\cdot;\vtheta^*)\right\}_{i=1}^M\right),
\end{equation}
where $\rspan\left\{ \phi_i(\cdot)\right\}_{i=1}^M=\{\sum_{i=1}^M a_i\phi_i(\cdot)|a_1,\cdots,a_M\in\sR\}$ and $\rdim(\cdot)$ returns the dimension of a linear function space. Then the model rank for any function $f^*\in\fF$ with model function space $\fF:=\{f(\cdot;\vtheta)|\vtheta\in\sR^M\}$ is defined as 
\begin{equation}
    R_{f_{\vtheta}}(f^*):=\min_{\vtheta'\in\vTheta_{f^*}}R_{f_{\vtheta}}(\vtheta'),
\end{equation}
where the target set $\vTheta_{f^*}:=\{\vtheta|f(\cdot;\vtheta)=f^*;\vtheta\in\sR^M\}$.
\end{definition*}

Given a differentiable model $f_{\vtheta}$, the optimistic estimate is obtained by the following two steps: estimate $R_{\vf_{\vtheta}}(\vtheta)$ for $\vtheta\in\sR^M$; and estimate $R_{\vf_{\vtheta}}(f)$ for $f\in\fF$. The difficulty of optimistic estimate depends on the complexity of model architecture. In the following subsections, we present details of optimistic estimates for deep models, matrix factorization models and two-layer tanh-NNs with various architectures with an upper bound for general DNNs.

\subsection{Deep model\label{appsec:DM}}
$f_{\vtheta}(\vx)=\sum_{j=1}^{d} \left(a_j^{[1]}\cdots a_j^{[L]}\right) x_j$ with $\vtheta=(\va_1,\cdots,\va_d)$, $\va_j=[a_j^{[1]},\cdots,a_j^{[L]}]^\TT\in\sR^L$, $L\geqslant 2$.

\textbf{Step 1: Estimate $R_{\vf_{\vtheta}}(\vtheta)$.}

For any parameter point $\vtheta$, the tangent space is 
$$
\operatorname{span} \left\{\dfrac{\partial f_{\vtheta}(\vx)}{\partial a_j^{[l]}}\right\}_{j\in [d], l\in[L]}=
\operatorname{span} \left\{\left(\prod_{s\in[L]\backslash \{l\}} a_j^{[s]}\right)x_j\right\}_{j\in [d], l\in[L]}=\bigoplus_{j=1}^{d}\operatorname{span} \left\{\left(\prod_{s\in[L]\backslash \{l\}} a_j^{[s]}\right)x_j\right\}_{l=1}^{L}.
$$
We observe that, for any $j\in[d]$, if $\va_j$ has more than one zeros, then $\operatorname{span} \left\{\left(\prod_{s\in[L]\backslash \{l\}} a_j^{[s]}\right)x_j\right\}_{l=1}^{L}=\operatorname{span}\{0(\cdot)\}$. Otherwise,  $\operatorname{span} \left\{\left(\prod_{s\in[L]\backslash \{l\}} a_j^{[s]}\right)x_j\right\}_{l=1}^{L}=\operatorname{span}\{x_j\}$. We define a function $g(\cdot)$ as $g(\vv)=0$ for vector $\vv$ having more than one zeros, and $g(\vv)=1$ for vector $\vv$ having no more than one zero. Then, we obtain
$$
\rdim\left(\operatorname{span} \left\{\left(\prod_{s\in[L]\backslash \{l\}} a_j^{[s]}\right)x_j\right\}_{l=1}^{L}\right)=g(\va_j)
$$
and 
$$
R_{\vf_{\vtheta}}(\vtheta)=\rdim\left\{\bigoplus_{j=1}^{d}\operatorname{span} \left\{\left(\prod_{s\in[L]\backslash \{l\}} a_j^{[s]}\right)x_j\right\}_{l=1}^{L}\right\}=\sum_{j=1}^{d}g(\va_j).
$$
\textbf{Step 2: Estimate $R_{\vf_{\vtheta}}(\vf)$.}

Given any target function $f^*\in\fF=\{\sum_{j=1}^{d}\alpha_j x_j|\alpha_1,\cdots,\alpha_d\in\sR\}$, we suppose it is $k$-sparse, i.e., it has $k$ nonzero coefficients. Then, for any $\vtheta\in\vTheta_{f^*}$, $\sum_{j=1}^{d}g(\va_j)\geqslant k$. Specifically, $\sum_{j=1}^{d}g(\va_j)$ attains $k$ when $\va_j=\vzero$ for all dimension $j$ with zero coefficients. Therefore, we have 
$$
R_{\vf_{\vtheta}}(\vf^*)=k.
$$

\subsection{Matrix factorization model\label{appsec:MF}}
$\boldsymbol{f}_{\boldsymbol{\theta}} = \mA \mB$ with $\vtheta=(\mA,\mB)$, $\mA = [a_{ij}]_{i, j=1}^{d}\in\sR^{d\times d}, \mB = [b_{ij}]_{i, j=1}^{d}\in\sR^{d\times d}$.

\textbf{Step 1: Estimate $R_{\vf_{\vtheta}}(\vtheta)$.}

% Stratify the parameter space into different model rank levels to obtain the rank hierarchy over the parameter space.
For any parameter point $\vtheta$, the tangent space is 
$
\operatorname{span} \big\{\mP^{ij}, \mQ^{ij}\big\}_{i, j=1}^{d},
$
where $\mP^{ij} = \dfrac{\partial\boldsymbol{f}_{\vtheta}}{\partial a_{ij}}=[\delta_{ki}b_{jl}]_{k,l=1}^{d}$, $\mQ^{ij} = \dfrac{\partial\boldsymbol{f}_{\vtheta}}{\partial b_{ij}}=[a_{ki}\delta_{jl}]_{k,l=1}^{d}$. Here $\delta_ij$ is the Kronecker delta.  By vectorizing $\mP^{ij}_{d\times d}$ and $\mQ^{ij}_{d\times d}$, we get $$\mathrm{vec}(\mP^{ij}) = [P^{ij}_{11}, \cdots, P^{ij}_{1d}, P^{ij}_{21}, \cdots, P^{ij}_{2d}, \cdots, P^{ij}_{d1}, \cdots, P^{ij}_{dd}]^{\TT}\in \sR^{d^2},$$
$$
\mathrm{vec}(\mQ^{ij}) = [Q^{ij}_{11}, \cdots, Q^{ij}_{1d}, Q^{ij}_{21}, \cdots, Q^{ij}_{2d}, \cdots, Q^{ij}_{d1}, \cdots, Q^{ij}_{dd}]^{\TT}\in \sR^{d^2}.
$$ 
Then, we have $$R_{\vf_{\vtheta}}(\vtheta)=\operatorname{dim}\left(\operatorname{span} \big\{\mP^{ij}, \mQ^{ij}\big\}_{i, j=1}^{d}\right)=\operatorname{dim}\left(\operatorname{span} \big\{\mathrm{vec}(\mP^{ij}), \mathrm{vec}(\mQ^{ij})\big\}_{i, j=1}^{d}\right)=\operatorname{rank}(\mGamma),
$$
where
$$ \mGamma = \left[
  \begin{matrix}
  \mathrm{vec}(\mQ^{11}), & \cdots,  \mathrm{vec}(\mQ^{dd}), & \mathrm{vec}(\mP^{11}), & \cdots, & \mathrm{vec}(\mP^{dd})
  \end{matrix}  \right]^{\TT}
$$ is a $2d^2\times d^2$ matrix.
Using the Kronecker product $\otimes$ of matrices, we obtain
$$
\mGamma= \left[\begin{array}{cccc}
\mB & & & \\
& \mB & & \\
& & \ddots & \\
& & & \mB \\
a_{11} \mI & a_{21} \mI & \cdots & a_{d 1} \mI \\
a_{12} \mI & a_{22} \mI & \cdots & a_{d 2} \mI \\
\vdots & \vdots & \ddots & \vdots \\
a_{1 d} \mI & a_{2 d} \mI & \cdots& a_{d d} \mI
\end{array}\right] = \left[
 \begin{matrix}
   \mI \otimes \mB \\
   \mA^{\TT}\otimes  \mI \\
  \end{matrix}
  \right].
$$
Then,  
$$
\operatorname{rank}(\mGamma) = \operatorname{rank}(\mGamma^\TT\mGamma) = \operatorname{rank}(\mI \otimes \mB^\TT\mB+\mA\mA^\TT\otimes  \mI) = \operatorname{rank}(\mB^\TT\mB\oplus \mA\mA^\TT),
$$
where $\oplus$ indicates the Kronecker sum. Let $r_{\mA}:= \rrank(\mA), r_{\mB} := \rrank(\mB)$. Because the eigenvalues of $\mB^\TT\mB\oplus \mA\mA^\TT$ are the sums of all possible pairs of eigenvalues of $\mB^\TT\mB$ and $\mA\mA^\TT$, there are $(d-r_{\mA})(d-r_{\mB})$ zero eigenvalues. Therefore, we have 
\begin{equation}\label{eq:mf_rank_SI}
R_{\vf_{\vtheta}}(\vtheta)=\operatorname{rank}(\mB^\TT\mB\oplus \mA\mA^\TT)=d^2 - (d-r_{\mA})(d-r_{\mB}).
\end{equation}

\textbf{Step 2: Estimate $R_{\vf_{\vtheta}}(\vf)$.}

Given any target matrix $\vf^*\in\sR^{d\times d}$, let $r=\rrank(\vf^*)$. Because
\begin{equation*}
    \rrank(\mA\mB)\leq \min\{\rrank(\mA), \rrank(\mB)\},
\end{equation*}
any factorization $\vf^*=\mA^*\mB^*$ satisfies $\rrank(\mA^*)\geqslant r$ and $\rrank(\mB^*)\geqslant r$. By Eq. \eqref{eq:mf_rank_SI}, $R_{\vf_{\vtheta}}(\vtheta^*) \geqslant d^2-(d-r)^2=2rd-r^2$. This lower bound can be obtained using singular value decomposition $\vf^*=\mU\mSigma\mV^\TT$, and let $\vtheta^*=(\mA^*,\mB^*)$ with $\mA^*=\mU\mSigma^{\frac{1}{2}}$ and $\mB^*=\mSigma^{\frac{1}{2}}\mV^\TT$. Then, 
\begin{equation}\label{eq:mf_rank_f_SI}
R_{\vf_{\vtheta}}(\vf^*)=R_{\vf_{\vtheta}}(\vtheta^*)=2rd-r^2.
\end{equation}
% with  attains its lower bound $2rd-r^2$, thus $R_{\vf_{\vtheta}}(\vf^*)=2rd-r^2$. 
% By Eq. \eqref{eq:mf_rank_f}, the matrix factorization model possesses the rank levels $\{2rd-r^2|r\in[d]\}$ over its function space, each of which is occupied by $\{\vf\in\sR^{d\times d}|\rrank(\vf)=r\}$. 

\subsection{Two-layer fully-connected tanh-NN}
$f_{\boldsymbol{\theta}}(\boldsymbol{x}) = \sum_{i=1}^{m}a_i\tanh(\boldsymbol{w}_i^\mathrm{T} \boldsymbol{x})$, with $\vx=[x_1,\cdots,x_d]^\TT\in \sR^{d}$, $\boldsymbol{\theta} = (a_i\in\sR, \boldsymbol{w}_i\in\sR^d)_{i=1}^m$.

\textbf{Step 1: Estimate $R_{\mathrm{NN}_m}(\vtheta)$.}

For any parameter point $\vtheta$, the tangent space is 
\begin{align*}
&\operatorname{span} \left\{\dfrac{\partial f_{\vtheta}(\vx)}{\partial a_i},\dfrac{\partial f_{\vtheta}(\vx)}{\partial w_{i1}},\cdots,\dfrac{\partial f_{\vtheta}(\vx)}{\partial w_{id}}\right\}_{i=1}^{m} \\
=&
\rspan \left\{\tanh(\vw_i^\TT\vx),a_i\tanh'(\vw_i^\TT\vx)x_1,\cdots, a_i\tanh'(\vw_i^\TT\vx)x_d\right\}_{i=1}^{m}\\
=&  \rspan \left\{\tanh(\vw_i^\TT\vx)\right\}_{i=1}^{m} \bigoplus \rspan \left\{a_i\tanh'(\vw_i^\TT\vx)x_1\right\}_{i=1}^{m} \bigoplus\cdots \bigoplus\rspan \left\{a_i\tanh'(\vw_i^\TT\vx)x_d\right\}_{i=1}^{m}.
\end{align*}
Then
$$
\rdim\left(\operatorname{span} \left\{\dfrac{\partial f_{\vtheta}(\vx)}{\partial a_i},\dfrac{\partial f_{\vtheta}(\vx)}{\partial w_{i1}},\cdots,\dfrac{\partial f_{\vtheta}(\vx)}{\partial w_{id}}\right\}_{i=1}^{m}\right)=\rdim \left(\rspan \left\{\tanh(\vw_i^\TT\vx)\right\}_{i=1}^{m}\right)+\rdim \left(\rspan \left\{a_i\tanh'(\vw_i^\TT\vx)x_1\right\}_{i=1}^{m}\right)d
$$
Let $m_{\vw}=\rdim \left(\rspan \left\{\tanh(\vw_i^\TT\vx)\right\}_{i=1}^{m}\right)$ and $m_{a}=\rdim \left(\rspan \left\{a_i\tanh'(\vw_i^\TT\vx)x_1\right\}_{i=1}^{m}\right)$, we obtain
\begin{equation}
    R_{\mathrm{NN}_m}(\vtheta)=m_{\vw}+ m_{a}d.
\end{equation}

\textbf{Step 2: Estimate $R_{\mathrm{NN}_m}(\vf)$.}

Given any $f^*\in\fF^{\mathrm{NN}}_{m}$, we suppose $f^* \in \fF^{\mathrm{NN}}_k/\fF^{\mathrm{NN}}_{k-1}$, i.e., the intrinsic width $k(f^*)=k$. Then, for any $\vtheta\in\vTheta_{f^*}$, we have $m_{\vw}\geqslant k$ and $m_{a}\geqslant k$. Therefore, $R_{\mathrm{NN}_m}(\vtheta)\geqslant k(d+1)$. $R_{\mathrm{NN}_m}$ attains $k(d+1)$ for a $\vtheta^*\in\vTheta_{f^*}$ constructed as follows. Because $f^* \in \fF^{\mathrm{NN}}_k/\fF^{\mathrm{NN}}_{k-1}$, it can be expressed by a width-$k$ NN as $\sum_{i=1}^{k}a_i'\tanh(\boldsymbol{w}_i'^\TT \vx)$ satisfying (i) $a_i'\neq0$ and $\vw_i'\neq\vzero$ for all $i\in[k]$ and (ii) $\vw_i'\neq\pm\vw_j'$ for any pair $i,j\in[k]$ and $i\neq j$. Then 
\begin{equation}\label{eq:ebd_nn}
\vtheta^*=(a_1',\vw_1',\cdots,a_k',\vw_k', 0,\vzero,\cdots,0,\vzero)
\end{equation}
satisfies $R_{\mathrm{NN}_m}(\vtheta^*)=k(d+1)$. Therefore, we obtain
\begin{equation}
    R_{\mathrm{NN}_m}(f^*)=k(f^*)(d+1).
\end{equation}

\subsection{Two-layer convolutional tanh-NN}
\begin{equation*}
    f_{\vtheta} (\vx) = \sum_{i=1}^{m_{\mathrm{C}}} \sum_{j=1}^{d+1-s} a_{ij} \tanh\left(\sum_{\alpha=1}^{s} x_{j+s-\alpha} K_{i;\alpha}\right),
\end{equation*}
with $\vx=[x_1,\cdots,x_d]^\TT\in \sR^{d}$, $\boldsymbol{\theta} = (\va_i\in\sR^{d+1-s},\vK_i\in\sR^s)_{i=1}^{m_{\rC}}$.

\textbf{Step 1: Estimate $R_{\mathrm{CNN}_{m_\rC}}(\vtheta)$.}

For any parameter point $\vtheta$, the tangent space is 
\begin{align*}
&\operatorname{span} \left\{\dfrac{\partial f_{\vtheta}(\vx)}{\partial a_{i1}},\cdots, \dfrac{\partial f_{\vtheta}(\vx)}{\partial a_{i(d+1-s)}}, \dfrac{\partial f_{\vtheta}(\vx)}{\partial K_{i;1}},\cdots,\dfrac{\partial f_{\vtheta}(\vx)}{\partial K_{i;s}}\right\}_{i=1}^{m_\rC} \\
=&
\rspan \left\{\tanh\left(\sum_{\alpha=1}^{s} x_{1+s-\alpha} K_{i;\alpha}\right),\cdots,\tanh\left(\sum_{\alpha=1}^{s} x_{d+1-\alpha} K_{i;\alpha}\right),\right.\\
&\left.\sum_{j=1}^{d+1-s} a_{ij} \tanh'\left(\sum_{\alpha=1}^{s} x_{j+s-\alpha} K_{i;\alpha}\right)x_{j+s-1},\cdots, \sum_{j=1}^{d+1-s} a_{ij} \tanh'\left(\sum_{\alpha=1}^{s} x_{j+s-\alpha} K_{i;\alpha}\right)x_{j} \right\}_{i=1}^{m_\rC}\\
=&  \rspan \left\{\tanh\left(\sum_{\alpha=1}^{s} x_{1+s-\alpha} K_{i;\alpha}\right)\right\}_{i=1}^{m_\rC} \bigoplus\cdots \bigoplus 
\rspan \left\{\tanh\left(\sum_{\alpha=1}^{s} x_{d+1-\alpha} K_{i;\alpha}\right)\right\}_{i=1}^{m_\rC}
\bigoplus\\
&
\rspan \left\{\sum_{j=1}^{d+1-s} a_{ij} \tanh'\left(\sum_{\alpha=1}^{s} x_{j+s-\alpha} K_{i;\alpha}\right)x_{j+s-1}\right\}_{i=1}^{m_\rC} \bigoplus\cdots \bigoplus 
\rspan \left\{\sum_{j=1}^{d+1-s} a_{ij} \tanh'\left(\sum_{\alpha=1}^{s} x_{j+s-\alpha} K_{i;\alpha}\right)x_{j}\right\}_{i=1}^{m_\rC}.
\end{align*}
Then
\begin{align*}
&\rdim\left(\operatorname{span} \left\{\dfrac{\partial f_{\vtheta}(\vx)}{\partial a_{i1}},\cdots, \dfrac{\partial f_{\vtheta}(\vx)}{\partial a_{i(d+1-s)}}, \dfrac{\partial f_{\vtheta}(\vx)}{\partial K_{i;1}},\cdots,\dfrac{\partial f_{\vtheta}(\vx)}{\partial K_{i;s}}\right\}_{i=1}^{m_\rC}\right)\\
=&\rdim\left(\rspan \left\{\tanh\left(\sum_{\alpha=1}^{s} x_{1+s-\alpha} K_{i;\alpha}\right)\right\}_{i=1}^{m_\rC}\right)(d+1-s)+
\rdim \left(\rspan \left\{\sum_{j=1}^{d+1-s} a_{ij} \tanh'\left(\sum_{\alpha=1}^{s} x_{j+s-\alpha} K_{i;\alpha}\right)x_{j}\right\}_{i=1}^{m_\rC}\right)s
\end{align*}
Let $m_{\vw}^{\rC}=\rdim\left(\rspan \left\{\tanh\left(\sum_{\alpha=1}^{s} x_{1+s-\alpha} K_{i;\alpha}\right)\right\}_{i=1}^{m_\rC}\right)$ and $m_{\va}^{\rC}=\rdim \left(\rspan \left\{\sum_{j=1}^{d+1-s} a_{ij} \tanh'\left(\sum_{\alpha=1}^{s} x_{j+s-\alpha} K_{i;\alpha}\right)x_{j}\right\}_{i=1}^{m_\rC}\right),$ we obtain
\begin{equation}
    R_{\mathrm{CNN}_{m_\rC}}(\vtheta)=m_{\vK}^{\rC}(d+1-s)+ m_{\va}^{\rC}s.
\end{equation}

\textbf{Step 2: Estimate $R_{\mathrm{CNN}_{m_\rC}}(\vf)$.}

Given any $f^*\in\fF^{\mathrm{CNN}}_{m_\rC}$, we suppose $f^* \in \fF^{\mathrm{CNN}}_k/\fF^{\mathrm{CNN}}_{k-1}$, i.e., the intrinsic number of kernels $k_{\rC}(f^*)=k$. Then, for any $\vtheta\in\vTheta_{f^*}$, we have $m_{\vK}^{\rC}\geqslant k$ and $m_{\va}^{\rC}\geqslant k$. Therefore, $R_{\mathrm{CNN}_{m_\rC}}(\vtheta)\geqslant k(d+1)$. $R_{\mathrm{CNN}_{m_\rC}}$ attains $k(d+1)$ for a $\vtheta^*\in\vTheta_{f^*}$ constructed as follows. Because $f^* \in \fF^{\mathrm{CNN}}_k/\fF^{\mathrm{CNN}}_{k-1}$, it can be expressed by a width-$k$ CNN as $\sum_{i=1}^{k} \sum_{j=1}^{d+1-s} a_{ij}' \tanh\left(\sum_{\alpha=1}^{s} x_{j+s-\alpha} K_{i;\alpha}'\right)$ satisfying (i) $\va_i'\neq\vzero$ and $\vK_i'\neq\vzero$ for all $i\in[k]$ and (ii) $\vK_i'\neq\pm\vK_j'$ for any pair $i,j\in[k]$ and $i\neq j$. Then 
\begin{equation}\label{eq:ebd_cnn}
\vtheta^*=(\va_1',\vK_1',\cdots,\va_k',\vK_k', 0,\vzero,\cdots,0,\vzero)
\end{equation}
satisfies $R_{\mathrm{CNN}_{m_\rC}}(\vtheta^*)=k(d+1)$. Therefore, we obtain
\begin{equation}
    R_{\mathrm{CNN}_{m_\rC}}(f^*)=k_{\rC}(f^*)(d+1).
\end{equation}
Following a similar procedure, we can also obtain the optimistic sample size for two-layer tanh-CNNs with $2$-d convolution. Suppose the input dimension is $d\times d$ and the kernel size is $s\times s$ with stride $1$, the optimistic sample size
\begin{equation}
    R_{\mathrm{CNN}_{m_\rC}}(f^*)=k_{\rC}(f^*)(s^2+(d+1-s)^2).
\end{equation}

\subsection{Two-layer no-sharing convolutional tanh-NN}
\begin{equation*}
    f_{\vtheta} (\vx) = \sum_{i=1}^{m_{\mathrm{C}}} \sum_{j=1}^{d+1-s} a_{ij} \tanh\left(\sum_{\alpha=1}^{s} x_{j+s-\alpha} K_{ij;\alpha}\right),
\end{equation*}
with $\vx=[x_1,\cdots,x_d]^\TT\in \sR^{d}$, $\boldsymbol{\theta} = (\va_i\in\sR^{d+1-s},\vK_{ij}\in\sR^s)_{i=1}^{m_{\rC}}$.

\textbf{Step 1: Estimate $R_{\mathrm{CNN}_{m_\rC}^{\mathrm{NS}}}(\vtheta)$.}

For any parameter point $\vtheta$, the tangent space is 
\begin{align*}
&\rspan \left\{\dfrac{\partial f_{\vtheta}(\vx)}{\partial a_{ij}}, \dfrac{\partial f_{\vtheta}(\vx)}{\partial K_{ij;1}},\cdots,\dfrac{\partial f_{\vtheta}(\vx)}{\partial K_{ij;s}}\right\}_{i\in[m_\rC],j\in[d+1-s]} \\
=&\rspan \left\{\tanh\left(\sum_{\alpha=1}^{s} x_{j+s-\alpha} K_{ij;\alpha}\right), a_{ij} \tanh'\left(\sum_{\alpha=1}^{s} x_{j+s-\alpha} K_{ij;\alpha}\right)x_{j+s-1},\cdots,\right.\\
&\left.a_{ij} \tanh'\left(\sum_{\alpha=1}^{s} x_{j+s-\alpha} K_{ij;\alpha}\right)x_{j} \right\}_{i\in[m_\rC],j\in[d+1-s]}\\
=&  \rspan \left\{\tanh\left(\sum_{\alpha=1}^{s} x_{1+s-\alpha} K_{ij;\alpha}\right)\right\}_{i\in[m_\rC],j\in[d+1-s]} \bigoplus \\
&
\rspan \left\{ a_{ij} \tanh'\left(\sum_{\alpha=1}^{s} x_{j+s-\alpha} K_{ij;\alpha}\right)x_{j+s-1}\right\}_{i\in[m_\rC],j\in[d+1-s]} \bigoplus\cdots \bigoplus\\
&\rspan \left\{ a_{ij} \tanh'\left(\sum_{\alpha=1}^{s} x_{j+s-\alpha} K_{ij;\alpha}\right)x_{j}\right\}_{i\in[m_\rC],j\in[d+1-s]}.
\end{align*}
Let $$m_{\vK}=\rdim\left(\rspan \left\{\tanh\left(\sum_{\alpha=1}^{s} x_{1+s-\alpha} K_{ij;\alpha}\right)\right\}_{i\in[m_\rC],j\in[d+1-s]}\right),$$ and $$m_{a}=\rdim \left(\rspan \left\{ a_{ij} \tanh'\left(\sum_{\alpha=1}^{s} x_{j+s-\alpha} K_{ij;\alpha}\right)x_{j}\right\}_{i\in[m_\rC],j\in[d+1-s]}\right),$$
we obtain
\begin{align}
&R_{\mathrm{CNN}_{m_\rC}^{\mathrm{NS}}}(\vtheta)=\rdim\left(\rspan \left\{\dfrac{\partial f_{\vtheta}(\vx)}{\partial a_{ij}}, \dfrac{\partial f_{\vtheta}(\vx)}{\partial K_{ij;1}},\cdots,\dfrac{\partial f_{\vtheta}(\vx)}{\partial K_{ij;s}}\right\}_{i\in[m_\rC],j\in[d+1-s]}\right)=m_{\vw}+m_{a}s.
\end{align}

\textbf{Step 2: Estimate $R_{\mathrm{CNN}_{m_\rC}^{\mathrm{NS}}}(\vf)$ for a subset of $\fF^{\mathrm{CNN-NS}}_{m_\rC}$.}

Given any $f^*\in\fF^{\mathrm{CNN-NS}}_{m_\rC}$, we suppose $f^* \in \fF^{\mathrm{CNN-NS}}_k/\fF^{\mathrm{CNN-NS}}_{k-1}$, i.e., the intrinsic number of kernels $k_{\rC}(f^*)=k$. Then $f^*$ can be expressed as $\sum_{i=1}^{k} \sum_{j=1}^{d+1-s} a_{ij}' \tanh\left(\sum_{\alpha=1}^{s} x_{j+s-\alpha} K_{ij;\alpha}'\right)$. If $f^*$ further satisfies (i) $a_{ij}'\neq 0$ and $\vK_{ij}'\neq\vzero$ for all $i\in[k],j\in[d+1-s]$ and (ii) $\tanh\left(\sum_{\alpha=1}^{s} x_{j+s-\alpha} K_{ij;\alpha}'\right)\neq\pm\tanh\left(\sum_{\alpha=1}^{s} x_{h+s-\alpha} K_{lh;\alpha}'\right)$ for all pairs $(i,j)\neq (l,h)$, then $m_{\vK}\geqslant k(d+1-s)$ and $m_{a}\geqslant k(d+1-s)$. Therefore, $R_{\mathrm{CNN}_{m_\rC}^{\mathrm{NS}}}(\vtheta)\geqslant k(d+1-s)(s+1)$. $R_{\mathrm{CNN}_{m_\rC}^{\mathrm{NS}}}$ attains $k(d+1)$ for 
\begin{equation}\label{eq:ebd_cnn_ns}
    \vtheta^*=(\va_{11}',\vK_{11}',\cdots,\va_{k(d+1-s)}',\vK_{k(d+1-s)}', 0,\vzero,\cdots,0,\vzero).
\end{equation}
Therefore, we obtain
\begin{equation}
    R_{\mathrm{CNN}_{m_\rC}^{\mathrm{NS}}}(f^*)=k(d+1-s)(s+1).
\end{equation}
Following a similar procedure, we can also obtain the optimistic sample size for two-layer no-sharing tanh-CNNs with $2$-d convolution. Suppose the input dimension is $d\times d$ and the kernel size is $s\times s$ with stride $1$, the optimistic sample size
\begin{equation}
    R_{\mathrm{CNN}_{m_\rC}^{\mathrm{NS}}}(f^*)=k(d+1-s)^2(s^2+1).
\end{equation}

\subsection{Deep fully-connected NNs\label{appsec:DNN}}

Estimating $R_{f_{\vtheta}}(f)$ for deep NNs with more than $2$ layers is challenging due to their complex composition structure. However, we note that it is not difficult to obtain a good upper bound estimate of $R_{f_{\vtheta}}(f)$. 

For a $L$-layer ($L\geq 2$) fully-connected NNs with a general differentiable activation function, $f_{\vtheta}(\cdot)$ is defined recursively: $\vf^{[0]}_{\vtheta}(\vx)=\vx$ for all $\vx\in\sR^d$; for $l\in[L-1]$, 
$\vf^{[l]}_{\vtheta}(\vx)=\sigma (\mW^{[l]} \vf^{[l-1]}_{\vtheta}(\vx)+\vb^{[l]})$; and the output function
\begin{equation}
    f_{\vtheta}(\vx)=f(\vx;\vtheta)=f^{[L]}_{\vtheta}(\vx)=\mW^{[L]} \vf^{[L-1]}_{\vtheta}(\vx)+b^{[L]}.
\end{equation} 
Here $W^{[l]}\in \sR^{m_l\times m_{l-1}}$, bias $b^{[l]}\in\sR^{m_{l}}$ for $l\in[L]$. $m_l$ is the number of neurons in the $l$-th layer with $m_0=d$ and $m_L=1$. The parameters
\begin{equation} \vtheta=\Big(\vtheta|_1,\cdots,\vtheta|_L\Big)=\Big(\mW^{[1]},\vb^{[1]},\ldots,\mW^{[L]},\vb^{[L]}\Big).
\end{equation}
where $\vtheta|_{l}=\Big(\mW^{[l]},\vb^{[l]}\Big)$. The total number of parameters $M=\sum_{l=0}^{L-1}(m_l+1) m_{l+1}$. For any $f^*\in\fF_{\{m_l'\}_{l=1}^{l-1}}^{\mathrm{NN}}\subseteq\fF^{\mathrm{NN}}_{\{m_l\}_{l=1}^{l-1}}$ with $m'_l\leqslant m_l$ for $l\in [L-1]$, we will show that the following upper bounds can be obtained
\begin{equation}
    R_{\{m_l\}_{l=1}^{l-1}}^{\mathrm{NN}}(f^*)\leqslant R_{\{m_l'\}_{l=1}^{l-1}}^{\mathrm{NN}}(f^*) \leqslant M'.
\end{equation}
Suppose $f^*=f_{\vtheta'}$, where $\vtheta'|_l=\left(\vW'^{[l]}\in\sR^{m'_l\times m'_{l-1}},\vb'^{[l]}\in\sR^{m_l'}\right)$ and $R_{\{m'_l\}_{l=1}^{l-1}}^{\mathrm{NN}}(f^*)=R_{\{m'_l\}_{l=1}^{l-1}}^{\mathrm{NN}}(\vtheta')$. We construct $\vtheta^*=\Big(\vtheta^*|_1,\cdots,\vtheta^*|_L\Big)=\fT_{\rnull}(\vtheta')$ as
\begin{equation}\label{eq:ebd_dnn}
\vtheta^*|_l=\fT_{\rnull}(\vtheta')|_l=\left(\left[
\begin{matrix}
    \vW'^{[l]} & \mzero_{m'_l\times(m_{l-1}-m'_{l-1})}\\
    \mzero_{(m_{l}-m'_{l})\times m'_{l-1}} & \mzero_{(m_{l}-m'_{l})\times(m_l-m'_l)}
\end{matrix}
\right],\left[
\begin{matrix}
    \vb'^{[l]} \\
    \mzero_{(m_{l}-m'_{l})}
\end{matrix}
\right]
\right)
\end{equation}
for all $l\in[L]$. Remark that $\fT_{\rnull}$ is a null embedding operator with output preserving and criticality preserving properties \cite{zhang2022embedding}. Then, we have
$$R_{\{m_l\}_{l=1}^{l-1}}^{\mathrm{NN}}(f^*)\leqslant R_{\{m_l\}_{l=1}^{L-1}}^{\mathrm{NN}}(\fT_{\rnull}(\vtheta')) = R_{\{m_l'\}_{l=1}^{L-1}}^{\mathrm{NN}}(\vtheta')=R_{\{m_l'\}_{l=1}^{L-1}}^{\mathrm{NN}}(f^*) \leqslant M'.$$
Above upper bound estimate shows the free expressiveness property in width for general deep NNs that the optimistic fitting performance never deteriorates as the NN becomes wider. We elaborate on this property in the main text.

% \subsection{Free expressiveness in width and the embedding principle}
\subsection{Remark on embedding the parameters of a narrow NN to a wider NN}
A key step in above optimistic estimates is to embed any $\vtheta'\in\sR^{M'}$ of a narrower NN to $\vtheta^*\in\sR^{M}$ 
of a wide NN such that $f_{\vtheta^*}=f_{\vtheta'}$ and 
$R_{\mathrm{NN}_\rwide}(\vtheta^*)=R_{\mathrm{NN}_\rnarr}(\vtheta^*)$. We remark that all critical embedding operators $\fT_\rc:\sR^{M'}\to\sR^{M}$ proposed in Refs. \cite{zhang2021embedding,zhang2022embedding} satisfy the above requirements. 
In particular, Eq. \eqref{eq:ebd_nn}, Eq. \eqref{eq:ebd_cnn}, 
 Eq. \eqref{eq:ebd_cnn_ns} and Eq. \eqref{eq:ebd_dnn} are essentially the null embedding in Ref. \cite{zhang2022embedding}. These critical embedding operators are initially proposed to study the embedding principle of the loss landscape of DNNs, i.e., the loss landscape of a wide NN inherits all critical points from that of the narrower NNs \cite{zhang2021embedding,zhang2022embedding}. Other studies about this embedding relation can be found in Refs. \cite{fukumizu2000local,fukumizu2019semi,csimcsek2021geometry}. It is clear from the above estimates that these critical embedding operators are important tools for the optimistic estimates of general DNNs. We will elaborate on the intimate relation between the optimistic estimate and the embedding principle in subsequent works.

\section{Details of experiments}\label{appsec:exp}
In Fig. 1, model parameters are initialized with a normal distribution with mean $0$ and variance $10^{-8}$, and trained with full-batch GD with a learning rate of $0.01$. The training ends when the training error is less than $10^{-9}$. 

In Fig. 2, model parameters are initialized with a normal distribution with mean $0$ and variance $10^{-8}$, and trained with full-batch GD with a learning rate of $0.05$. The training ends when the training error is less than $10^{-9}$. The target matrices we use are as follows:
\begin{align*}
&\mM_{1}^*=\left[\begin{array}{cccc}
1 &0.3 &0.7 & -0.4   \\
2 &   0.6 & 1.4 &-0.8   \\
4 &   1.2&  2.8& -1.6   \\
 7 &  2.1 & 4.9 & -2.8   \\
\end{array}\right],    
&&\mM_{2}^*=\left[\begin{array}{cccc}
4 &  0.6 & 1.8 & 0.8   \\
6 &  0.9 & 2.7 & 1.2   \\
8 &  1.2 & 3.6 & 1.6   \\
18  & 2.7 & 8.1 & 3.6   \\
\end{array}\right],\\
&\mM_{3}^*=\left[\begin{array}{cccc}
-1.8 & 2.4 & 7.7 & -5.3   \\
0.4 & 1.8 & 5.4 & -3.6   \\
3.2 & 1.8 & 4.8 & -3.   \\
6.6 & 2.4 & 5.9 & -3.5   \\
\end{array}\right],  
&& \mM_{4}^*=\left[\begin{array}{cccc}
7.6 & 3.3 & 19.8 & -7.3   \\
7.6 & 2.1 & 10.7 & -2.4   \\
8.8 & 1.8 & 7.6 & -0.2   \\
19.2 & 3.6 & 14.1 &  0.9   \\
\end{array}\right],\\   
&\mM_{5}^*=\left[\begin{array}{cccc}
-1.8 & 2.4 & 7.7 & -5.3  \\
 0.4 & 1.8 & 5.4 & -3.6  \\
3.2 & 1.8 & 4.8 & -3   \\
6.6 & 2.4 & 5.9 & -3.5   \\
\end{array}\right],
&&\mM_{6}^*=\left[\begin{array}{cccc}
8.5 & 9.3 & 22.5 & -6.1  \\
8.2 & 6.1 & 12.5 & -1.6   \\
11.5 & 19.8 & 15.7 &  3.4   \\
20.4 & 11.6 & 17.7 & 2.5   \\
\end{array}\right],\\
&\mM_{7}^*=\left[\begin{array}{cccc}
3.6 & -1.2 & 8.1 & -3.5   \\
8.1 & -3.5 & 3.6 & -1.2   \\
 9.1 & -1.7 & 11.4 & -0.6   \\
11.4 & -0.6 & 9.1 & -1.7   \\
\end{array}\right],
&&\mM_{8}^*=\left[\begin{array}{cccc}
12.1 & 17.3 & 24.1 & -4.9   \\
16.3 & 24.1 & 16.1 & 1.1   \\
14.2 & 25.8 & 16.9 & 4.3   \\
22.2 & 15.6 & 18.5 & 3.1   \\
\end{array}\right].
\end{align*}

In Fig. 3b, We use $\mM_{1}^*$ above as the target matrix.  For all experiments, the network is initialized by a normal distribution with mean $0$ and variance $\sigma^2$, and trained with full-batch GD with a learning rate $0.05$. 

In Fig. 4, we use fully-connected NNs or CNNs with bias term $b$, e.g., $f_{\vtheta}=\sum_{i=1}^{m}a_i\tanh(\vw_i^T\vx+b_i)$, to fit the target function. Therefore, a $m$-kernel CNN has $7m$ parameters, and the corresponding width-$3m$ fully-connected NN has $21m$ parameters. Fully-connected NNs and CNNs with $m=1$, $3$, $10$, $34$, or $100$ are used for fitting. Network parameters are initialized by a normal distribution with mean $0$ and variance $10^{-20}$, and trained by full-batch gradient descent. The learning rate for the experiments in each setup is fine-tuned from $0.05$ to $0.5$ for a better generalization performance. For the training dataset and the test dataset, we construct the input data through the standard normal distribution and obtain the output values from the target function. The size of the training dataset varies whereas the size of the test dataset is fixed to $1000$.

\end{document}